\documentclass[12pt]{article}

\usepackage{newtxtext,newtxmath}

\usepackage{graphicx}

\usepackage[letterpaper,margin=1in]{geometry}

\renewenvironment{abstract}
	{\quotation}
	{\endquotation}

\date{}

\makeatletter
\renewcommand{\fnum@figure}{\textbf{Figure \thefigure}}
\renewcommand{\fnum@table}{\textbf{Table \thetable}}
\makeatother

\usepackage{scicite}

\usepackage{url}

\def\scititle{Success in Humanoid Reinforcement Learning under Partial Observation}
\title{\bfseries \boldmath \scititle}

\author{
	Wuhao Wang, 
	Zhiyong Chen$^{\ast}$\\
    \small School of Engineering, The University of Newcastle, Callaghan, NSW 2308, Australia.\and
	\small$^\ast$Corresponding author. Email: zhiyong.chen@newcastle.edu.au\and
}

\begin{document} 

\maketitle

\begin{abstract} 
Reinforcement learning has been widely applied to robotic control, but effective policy learning under partial observability remains a major challenge, especially in high-dimensional tasks like humanoid locomotion. To date, no prior work has demonstrated stable training of humanoid policies with incomplete state information in the benchmark Gymnasium Humanoid-v4 environment. The objective in this environment is to walk forward as fast as possible without falling, with rewards provided for staying upright and moving forward, and penalties incurred for excessive actions and external contact forces.
This research presents the first successful instance of learning under partial observability in this environment. The learned policy achieves performance comparable to state-of-the-art results with full state access, despite using only one-third to two-thirds of the original states. Moreover, the policy exhibits adaptability to robot properties, such as variations in body part masses.
The key to this success is a novel history encoder that processes a fixed-length sequence of past observations in parallel. Integrated into a standard model-free algorithm, the encoder enables performance on par with fully observed baselines. We hypothesize that it reconstructs essential contextual information from recent observations, thereby enabling robust decision-making.
 

\end{abstract}

\noindent
\subsection*{Introduction}

The long-standing vision of robotics is to develop general-purpose machines capable of operating in diverse, unstructured environments. Humanoid robots with their anthropomorphic design are particularly well-suited for tasks requiring physical dexterity, adaptability, and autonomy in spaces built for humans. Recent advances in hardware and motion planning have enabled such robots to climb stairs~\cite{Cipriano2023Humanoid} and even fold laundry~\cite{Estevez2020Enabling}. Yet, controlling humanoids remains a significant challenge due to their high dimensionality and complex dynamics. While reinforcement learning (RL) has shown promise in fully observed simulation settings~\cite{Silva2021Deep, RohanP2022Learning}, its effectiveness under partial observability, particularly in high-degree-of-freedom (DoF) systems like humanoids, remains largely unproven.

Partial observability is a natural characteristic of real-world robotic systems, arising from incomplete sensing, measurement noise, or limited bandwidth. Unlike fully observable Markov decision processes (MDPs)~\cite{Bellman1957DP}, such scenarios require modeling as partially observable Markov decision processes (POMDPs)~\cite{astrom1965optimal}, where the agent must infer latent states from incomplete observations, requiring additional memory or belief inference mechanisms~\cite{Liu2022When, Wang2023Learning, Chen2022Flow, Wei2023Set}. Prior approaches commonly adopt recurrent neural networks (RNNs), such as Long-Short-Term Memory (LSTM)~\cite{hochreiter1997long, Meng2021Memory, Ni2022Recurrent, Zhao2023ODE, Arcieri2024POMDP, Lemmel2025Real} or recent architectures like Mamba~\cite{Gu2023Mamba, Dao2024Transformers, Ota2024Decision, Rimon2024World}, to encode temporal dependencies. While these memory-based methods have demonstrated promise in simpler POMDP benchmarks, they have yet to show success in highly complex environments such as humanoid.

In this work, we introduce a novel model-free RL method that enhances exploration by encoding fixed-length histories of past observations in parallel, rather than relying on the sequential logic used in memory-based approaches. By treating each time step in the history as equally important, our network allows the agent to flexibly combine information across the observation window. We hypothesize that such neural networks can infer complete latent states from successive observation sequences.


We evaluate our method in the Gymnasium Humanoid-v4 environment, a high-dimensional continuous control benchmark with complex dynamics and large state-action spaces. Prior work on POMDPs has primarily focused on simpler environments, and applying reinforcement learning under partial observability in this setting remains a significant challenge. Notably, to date, no existing approach has demonstrated stable training of humanoid policies using incomplete state information.

We compare our method against two recent state-of-the-art memory-based approaches~\cite{Ni2022Recurrent, Zhao2023ODE}, which have demonstrated success on general POMDP benchmarks. While these baselines struggle to learn effective policies in the humanoid task under partial observation, our approach enables agents to achieve policy performance comparable to training under full observability. These results suggest that the proposed temporal encoder effectively reconstructs essential latent state information from recent observations, and that strategic context aggregation can even outperform full observation in certain settings. 
The findings highlight a promising direction for reinforcement learning in high-dimensional, partially observed robotic control tasks, and they motivate further research into sensor-efficient, model-free policy learning for complex domains such as humanoid locomotion.


\subsection*{Results}

We conducted all experiments in the Mujoco Humanoid-v4 environment of Gymnasium~\cite{Towers2023Gym}, a high-dimensional continuous control task featuring a 348-dimensional state space and a 17-dimensional action space. The agent controls a simulated bipedal humanoid robot by applying continuous joint torques to maintain balance and achieve locomotion.

To simulate partial observability, we construct environments by selectively removing specific components from the original state vector. We categorize the full 348-dimensional state into four semantic attributes: position (position values of body parts, 22 dimensions), velocity (velocities of body parts and center-of-mass-based velocities, 101 dimensions), mass/inertia (mass and inertia of the rigid body parts, 130 dimensions), and force (actuator force at each joint and external forces on the body parts, 95 dimensions).

Removing position information breaks the causal link between observations and actions, making it difficult for the agent to learn meaningful control policies. Therefore, we retain joint positions in all settings and selectively remove other attributes, including 
\underline{V}elocity, \underline{M}ass/inertia, and \underline{F}orce, as well as their combinations. Table~\ref{tab:obs_config} lists the resulting observation configurations and reports the percentage of retained dimensions relative to the full observation space of 348 in each case.
Removing one attribute retains approximately two-thirds of the original state dimensions, while removing two attributes reduces the observation space to about one-third of its original size.

\begin{table}[h]
\centering
\caption{\textbf{Observation dimension for each partial observability configuration}}
\label{tab:obs_config}
\resizebox{\textwidth}{!}
{%
\begin{tabular}{lcccccc}
\hline
Setting       & Remove V & Remove M & Remove F  & Remove VM & Remove VF & Remove MF \\
\hline
Obs. Dim. (\%) & 247 (71\%) & 218 (63\%) & 253 (73\%) & 117 (34\%) & 152 (44\%) & 123 (35\%)  \\
\hline
\end{tabular} 
}
\end{table}

We compare our method against two recent state-of-the-art approaches designed for POMDPs: (1) \underline{RMF}, a recurrent model-free RL baseline using LSTM networks~\cite{Ni2022Recurrent}, and (2) \underline{ODERMF}, an ODE-based recurrent architecture~\cite{Zhao2023ODE}. All models are implemented within a unified training framework, using matched network capacity and consistent hyperparameters to ensure fair comparison. Each method is trained for 1 million or 3 million gradient steps across various partially observable settings, using five random seeds. Performance is evaluated using episodic return curves averaged over seeds. 
As a benchmark for training performance, we include \underline{TD3 with full state observations}. In the experiments, most hyperparameters follow the original TD3 configuration. Further details will be provided in a separate report that discusses the architecture and mechanisms of our method.
 
All experiments were conducted on NVIDIA Tesla V100 (Volta architecture) GPU compute nodes provided by the Australian National Computational Infrastructure (NCI). Each experiment used a single NVIDIA V100 GPU (32 GB HBM2 memory) and 12 CPU cores of Intel Xeon Gold 6148 processors (2.4 GHz). The average system memory usage was 4.36 GB, and GPU memory utilization remained at 464 MB. Running 3 million training steps required approximately 47.7 hours per experiment. Under our implementation, GPU utilization was around 10\%, primarily because environment interactions in the Humanoid simulation are CPU-bound and constitute the main computational bottleneck.

\subsubsection*{Removal of One Observation Attribute}

\begin{figure}[t] 
	\centering
	\includegraphics[width=1\textwidth]{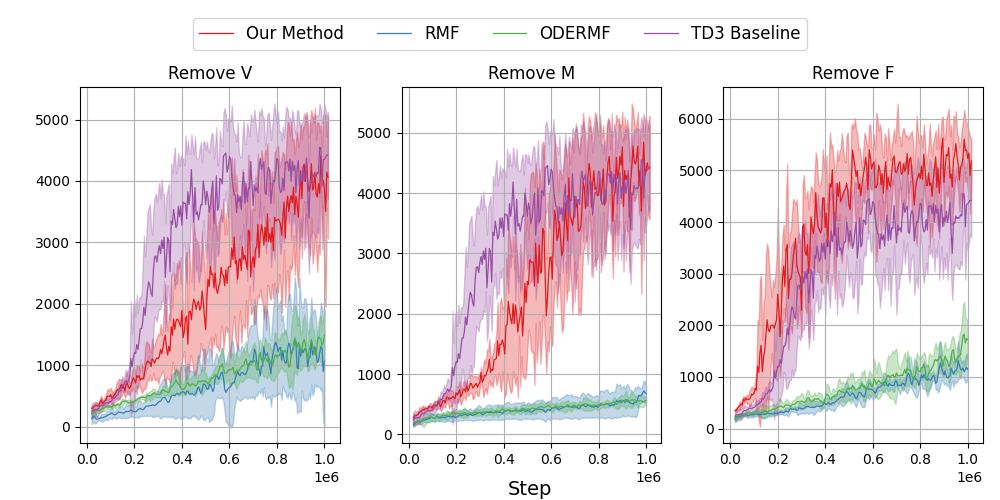} 
	\caption{
\textbf{Training performance of three POMDP methods with one observation attribute removed, compared to the TD3 baseline with full state observations.} 
}
	\label{fig:main_result} 
\end{figure}

Figure~\ref{fig:main_result} summarizes the training performance under three partial observability configurations: Remove V, Remove M, and Remove F. We also include the numeric records in Table~\ref{tab:main_res}.
It includes the average maximum reward and reward during the last 25\% of training for each partial observability configuration.
The last 25\% reward refers to the evaluation reward collected during the final 25\% of training (e.g., from 0.75M to 1M steps in a 1M-step training run).

The RMF and ODERMF algorithms are not considered successful, as their performance was substantially lower, rendering them incomparable to the TD3 baseline and highlighting their limitations in high-dimensional control. 
In contrast, our method consistently outperforms both RMF and ODERMF, achieving performance comparable to the TD3 baseline trained with full observations.

Our method remains robust across all partial observability configurations. Among the observation attributes, velocity is the most critical for performance, while mass and inertia notably influence convergence speed. In contrast, force information appears to be the least essential. Remarkably, in the Remove M and Remove F configurations, our method even surpasses the TD3 baseline trained with full observations. This suggests that the encoder effectively reconstructs missing information and exploits redundancy. To further demonstrate the potential of our algorithm, we extended the training to 3 million gradient steps beyond the initial 1 million-step comparison.

\begin{table}[h]
\centering
\caption{\textbf{Average maximum reward / reward during the last 25\% of training with removal of one observation attribute.}}
\label{tab:main_res}
\resizebox{\textwidth}{!}
{%
\begin{tabular}{lccc}
\hline
Method & Remove V & Remove M & Remove F\\
\hline
RMF (1M) &
1428.0 / 1189.0 ± 144.6 &
721.1 / 528.4 ± 37.1 &
1316.6 / 1040.9 ± 91.1\\
ODERMF (1M) &
1461.0 / 1230.4 ± 93.8 &
580.7 / 525.0 ± 26.2 &
1827.6 / 1284.0 ± 98.7\\
Ours (1M) &
4410.3 / 3658.2 ± 202.7 &
4844.6 / 4247.8 ± 230.9 &
5625.3 / 5017.3 ± 178.8\\
Ours (3M) &
5137.1 / 4573.4 ± 209.1 &
5037.3 / 4387.3 ± 214.2 &
6680.8 / 5776.4 ± 238.1\\
\hline 
TD3 (1M) &
\multicolumn{3}{c}{4547.2 / 4092.2 $\pm$133.2}\\
\hline
\end{tabular}
}
\end{table}

\subsubsection*{Removal of Two Observation Attributes}

\begin{figure}[t] 
	\centering
	\includegraphics[width=0.7\textwidth]{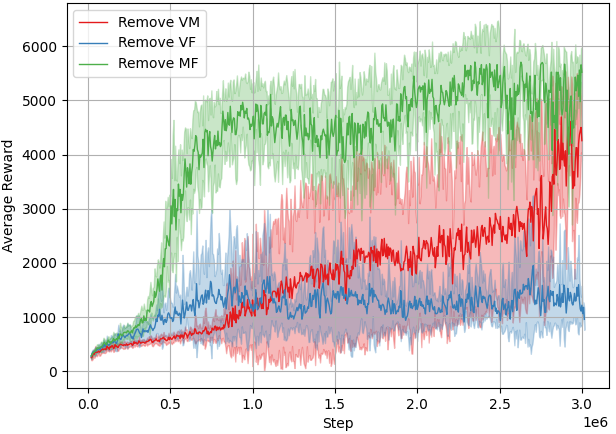} 
	\caption{
\textbf{Training performance of our method with two observation attributes removed.}
}
	\label{fig:remove_double} 
\end{figure}

The results for the three configurations with two observation attributes removed are reported in Fig.~\ref{fig:remove_double}, highlighting the impact of missing combinations of observation attributes on training performance. The corresponding numerical results are provided in Table~\ref{tab:remove_two_attr}.
As the RMF and ODERMF algorithms fail when a single observation attribute is removed, they are not included in this comparison. The TD3 baseline with full state observations remains the same and is not repeated here.
 
When velocity and mass/inertia are removed (Remove VM), reconstructing dynamic state information from position and force alone becomes significantly more challenging, resulting in much slower convergence. However, given sufficient training time, the model can still partially recover the missing information and achieve meaningful performance. Similarly, removing velocity and force (Remove VF) hinders the estimation of state transitions, as position combined with mass and inertia does not provide sufficient information to infer the full motion states. In contrast, when mass/inertia and force are removed (Remove MF), velocity remains observable and offers sufficient cues to model movement, enabling successful learning. In this case, our method still converges relatively quickly and eventually surpasses the performance of TD3 trained with full observations. These results suggest that our method performs reliably when the available partial observations contain sufficient information to reconstruct the underlying dynamics.


\begin{table}[h]
\centering
\caption{\textbf{Average maximum reward / reward during the last 25\% of training with removal of two observation attributes.}}
\label{tab:remove_two_attr}
{%
\begin{tabular}{lccc}
\hline
Method & Remove VM & Remove VF & Remove MF\\
\hline
Ours (1M) &
1339.0 / 974.6 ± 535.3 &
1781.5 / 1382.1 ± 730.9 &
4964.9 / 4537.3 ± 651.0 \\
Ours (3M) &
4691.7 / 2978.3 ± 1310.6 &
1951.5 / 1308.7 ± 505.8 &
5707.7 / 5175.7 ± 691.7 \\
\hline
\end{tabular}
}
\end{table}

\subsubsection*{Learning under Varying Body Masses}

As successful training can still be achieved when mass/inertia and force inputs are removed (Remove MF), it suggests that the learned policy is adaptive to variations in body mass and, by extension, inertia. The 3D bipedal robot body consists of a torso and pelvis, with a pair of legs and arms, and additional components. Each leg comprises three segments (thigh, shin, and foot), while each arm includes two segments (upper arm and forearm).

\begin{figure}[t] 
	\centering
	\includegraphics[width=0.7\textwidth]{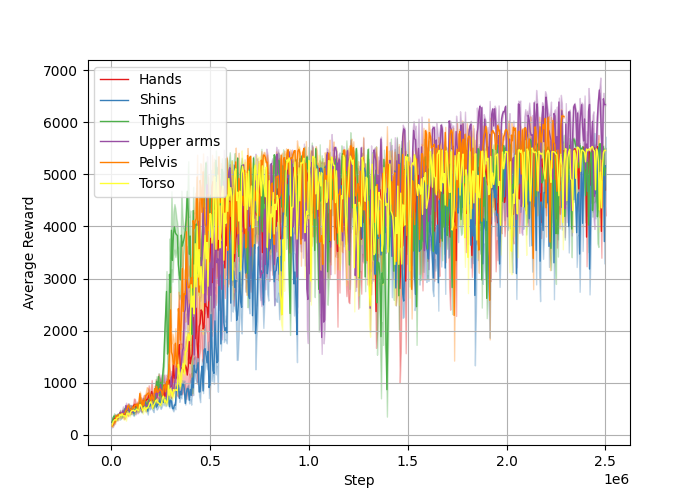}  
	\caption{
\textbf{Training performance of our method with body part masses varied every 10,000 steps.}
}
	\label{fig:mass_vary} 
\end{figure}

Figure~\ref{fig:mass_vary} summarizes the training performance when the mass of each body part, specifically, the two hands, two shins, two thighs, two upper arms, pelvis, and torso, is varied every 10,000 steps. The variation range is between 50\% and 100\% of the default mass values. These results demonstrate that the training process is adaptive to mass changes in all body parts.

We further evaluate one learned policy shown in Fig.~\ref{fig:mass_vary}, which was trained with the masses of the two upper arms varied. Table~\ref{tab:eveluation_mass} presents the results across various conditions, where the mass of each body part is independently varied by $\pm 50\%$. Each scenario is evaluated ten times, and the average performance is reported in the table. The results verify that the learned policy exhibits adaptability to mass variations across all body parts.

\begin{table}[h]
\centering
\caption{\textbf{Evaluation of a learned policy on the Humanoid with body part masses varied by $\pm 50\%$.}}
\label{tab:eveluation_mass}
{%
\begin{tabular}{cccccc}
\hline
 Hands & Shins & Thighs & Upper arms & Pelvis & Torso\\
\hline
  5542.5 & 6084.7 & 5492.9 & 6329.3 & 6697.8 & 5607.2\\
\hline
\end{tabular}
}
\end{table}


\subsection*{Discussion}

Our work demonstrates that model-free reinforcement learning with a novel parallel history encoding can successfully support training for humanoid locomotion under partial observability. Certain observation attributes, such as position and velocity, play a more critical role in inferring the full system state than others, such as mass/inertia and force. The proposed approach shows promise in reducing reliance on extensive sensor setups and enhancing training robustness in complex systems. We hope these results will inspire further investigation into lightweight temporal encoding strategies and their application to real-world robotic control tasks.

\bibliography{science_template} 
\bibliographystyle{sciencemag}

\newpage


\renewcommand{\thefigure}{S\arabic{figure}}
\renewcommand{\thetable}{S\arabic{table}}
\renewcommand{\theequation}{S\arabic{equation}}
\renewcommand{\thepage}{S\arabic{page}}
\setcounter{figure}{0}
\setcounter{table}{0}
\setcounter{equation}{0}
\setcounter{page}{1} 

\end{document}